\documentclass[conference]{IEEEtran}
\IEEEoverridecommandlockouts
\usepackage{cite}
\usepackage{amsmath,amssymb,amsfonts}
\usepackage{algorithmic}
\usepackage{graphicx}
\usepackage{textcomp}
\usepackage{xcolor}
\usepackage{booktabs}
\usepackage{makecell}
\usepackage{multirow}

\pdfoutput=1

\newcommand{\methodname}{{\tt{EFF-DVP}}}

\def\BibTeX{{\rm B\kern-.05em{\sc i\kern-.025em b}\kern-.08em
    T\kern-.1667em\lower.7ex\hbox{E}\kern-.125emX}}
\begin{document}

\title{Towards Group Fairness with Multiple Sensitive Attributes in Federated Foundation Models}

\author{Yuning Yang, Han Yu, Tianrun Gao, Xiaodong Xu, and Guangyu Wang}

\maketitle

\begin{abstract}

The deep integration of foundation models (FM) with federated learning (FL) enhances personalization and scalability for diverse downstream tasks, making it crucial in sensitive domains like healthcare. Achieving group fairness has become an increasingly prominent issue in the era of federated foundation models (FFMs), since biases in sensitive attributes might lead to inequitable treatment for under-represented demographic groups. Existing studies mostly focus on achieving fairness with respect to a single sensitive attribute. This renders them unable to provide clear interpretability of dependencies among multiple sensitive attributes which is required to achieve group fairness.
Our paper takes the first attempt towards a causal analysis of the relationship between group fairness across various sensitive attributes in the FFM. We extend the FFM structure to trade off multiple sensitive attributes simultaneously and quantify the causal effect behind the group fairness through causal discovery and inference. Extensive experiments validate its effectiveness, offering insights into interpretability towards building trustworthy and fair FFM systems.

\end{abstract}


\section{Introduction}
\label{sec:intro}

Federated learning (FL) is a paradigm of collaborative model training without exposing raw data, thereby enabling artificial intelligence (AI) model building with private data from sensitive domains such as healthcare \cite{mcmahan2017communication, kairouz2021advances, gao2023fedmbp, yang2024dense}. Recently, the integration of foundation models (FMs) with FL has further enhanced the personalization and generalization abilities of the resulting models, offering scalable solutions for various downstream tasks (e.g., medical imaging analysis, personalized diagnostics, predictive analytics) \cite{qu2022rethinking, yan2024buffalo}. In particular, parameter-efficient fine-tuning (PEFT) empowers FMs to adapt to domain-specific tasks and propels advances in communication and computation efficiency of federated foundation models (FFMs) training \cite{ren2024advances, gao2024prospect}.

\begin{figure}[t]
    \centering
    \includegraphics[width=1\linewidth]{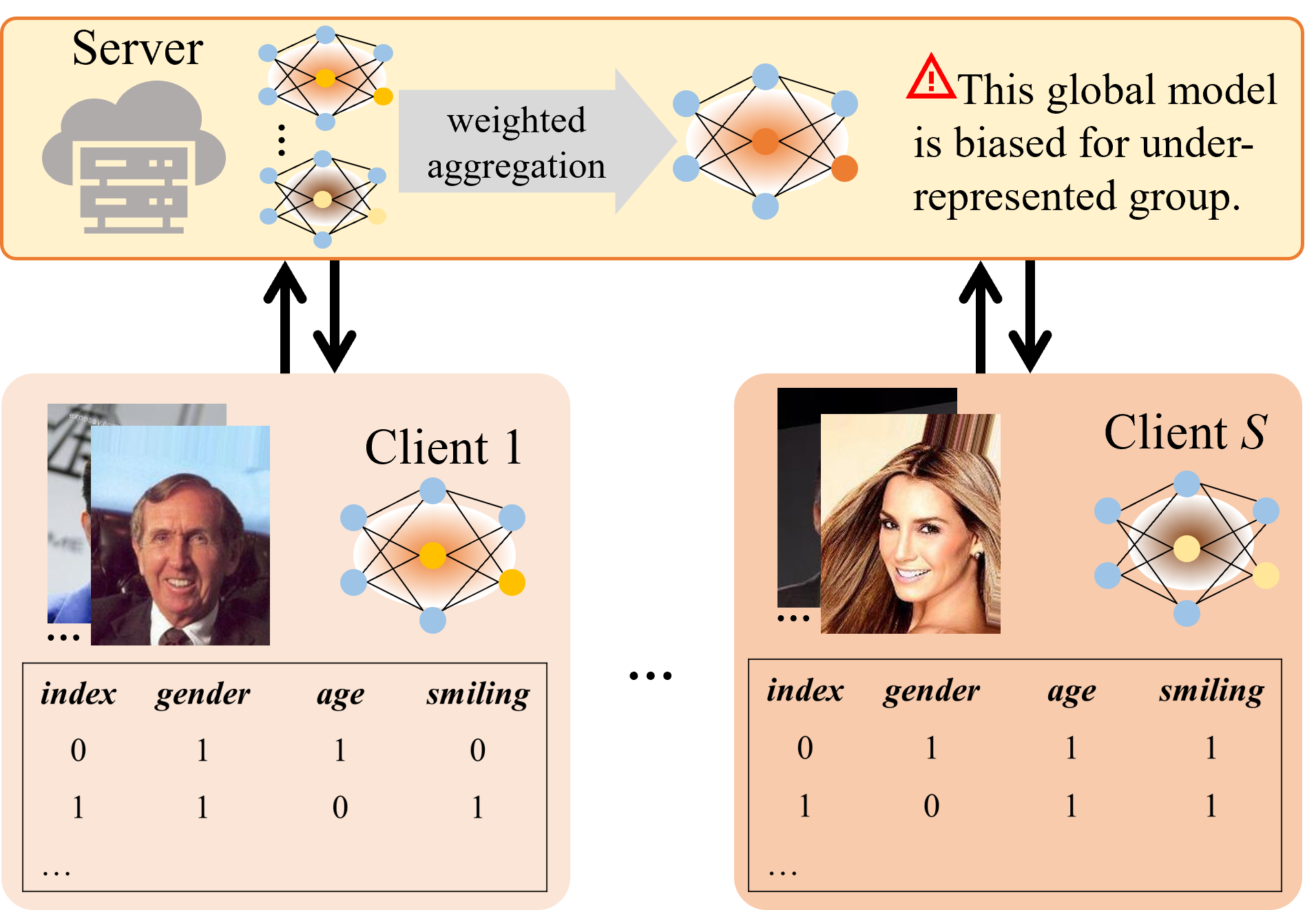}
    \caption{An example of the biased FL system. Gender: male (1) and female (0); Age: young (1) and old (0); Smiling: smile (1) and not smile (0). Client 1 is biased in the under-represented female group, while client $S$ is biased in the old group. By weighted averaging, the global model inherits the bias and behaves the unfairness for under-represented groups.}
    \label{fig:bias}
\end{figure}

As FFMs continue to gain prominence across various domains, ensuring the fairness of the resulting models has become critical for building trust \cite{tariq2023trustworthy}. Biases as a result of data quality, imbalanced data distributions and diverse local resources among FL clients can lead to inequitable performance of the resulting FFMs (Figure \ref{fig:bias}), thereby risking the sustainable development of FL ecosystems \cite{shi2023towards}. In many application domains, group fairness, which aims to ensure that individuals receive equitable algorithmic decisions regardless of their sensitive attributes (e.g., age, gender, race), plays a particularly pivotal role in fostering user trust in AI \cite{hosseini2023proportionally, zhang2024unified}. For example, based on the sensitive attribute of gender, male patients might be viewed as the unprotected group, while female patients as the protected group if they have been historically under-represented. Such disparities, if not handled properly, might lead FFMs to predict disproportionately higher risks for certain diseases for the under-represented group \cite{chen2023algorithmic}.

\begin{figure*}[t]
    \centering
    \includegraphics[width=1\linewidth]{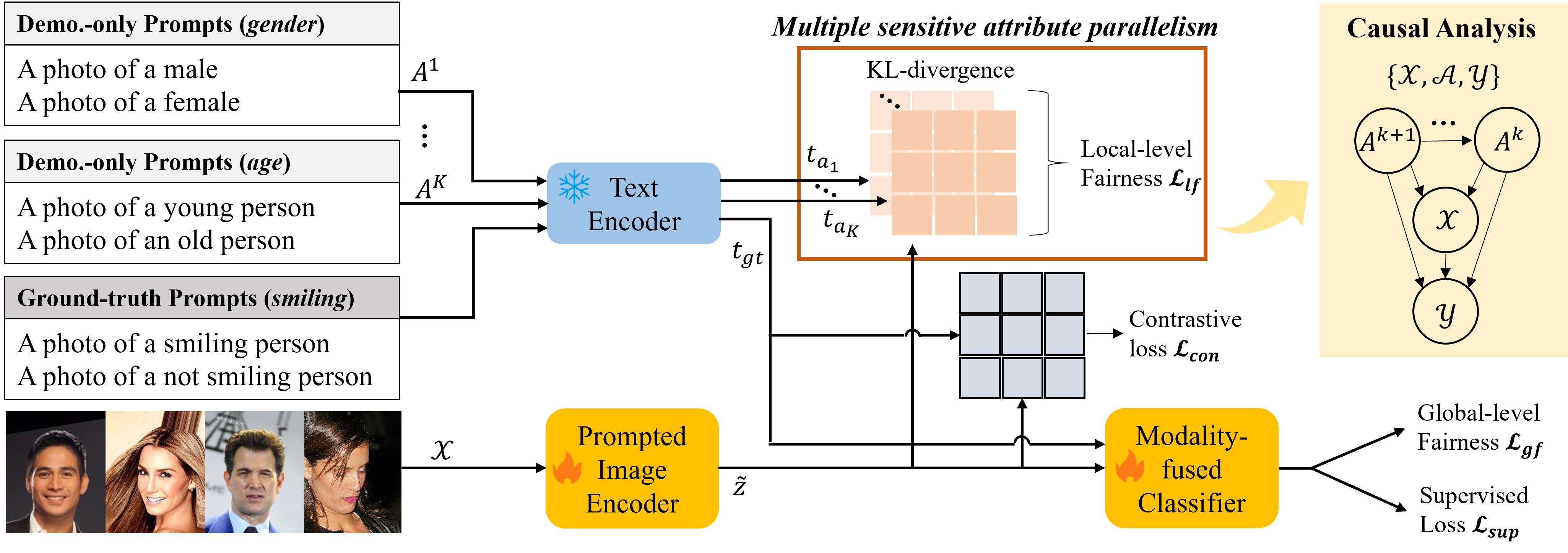}
    \caption{The local model overview of \methodname{}. Orange modules with the fire logo can be updated and shared with the server, while blue modules with the snow logo are fixed. Multiple sensitive attributes are converted to demographic-only prompts for text encoding and processed in parallel to trade off fairness. In addition, the causal analysis is observed simultaneously to reveal the dependency of sensitive attributes with group fairness.}
    \label{fig:overview}
\end{figure*}

FL group fairness has been studied extensively. Existing approaches can be divided into three categories: 1) aggregation-based, 2) reweighting-based, and 3) client selection-based methods. Aggregation-based methods leverage a proxy dataset on the server for calculating fairness metrics and alleviate the sensitive group biases from a global level without the need for direct client involvement \cite{kanaparthy2022fair, zeng2024fair}. Reweighting-based methods modify clients' model weights based on the group distributions to achieve equitable predictions \cite{du2021fairness, ezzeldin2023fairfed}. Client selection-based methods assess the contribution made by each client towards group fairness and give priority to those with higher contributions to join FL \cite{zhang2020fairfl, salazar2023fair}. 

Despite the advances, existing methods face a key limitation: they mostly consider group fairness regarding a single sensitive attribute. Studies have shown that mitigating bias regarding one sensitive attribute might damage the FL group fairness towards other sensitive attributes \cite{meerza2024glocalfair,djebrouni2024bias}. Although they attempt to solve this problem by incorporating fairness constraints into the training process, such optimization-based methods lack interpretability, making it challenging to understand the rationale behind an FFM’s fairness reasoning.

To address this challenge, we take the first step towards offering a structured approach to understand the interplay between FFM group fairness across multiple sensitive attributes via causal analysis. To this end, we extend the architecture of a typical FFM to make it capable of dynamically trading off multiple sensitive attributes simultaneously to achieve group fairness. Furthermore, causal analyses are incorporated to improve the interpretability of the relationship between sensitive attributes and group fairness. Specifically, causal discovery is conducted to construct a causal graph to reveal the causal relationships among attributes and labels \cite{guo2024fedcsl}. Causal Inference is conducted to verify the genuine causal relationships and quantitatively estimate the causal effects among variables \cite{li2024causal}. Through extensive experimental studies, we discovered that sensitive attributes which have strong causal effects with the labels are more likely to confound group unfairness. This insight paves the way for designing strategies for balancing fairness across multiple sensitive attributes and promoting equitable outcomes in FFM systems.

\section{Related Works}
\label{sec:rw}

The challenge of fairness is becoming increasingly prominent in the era of FFM due to the amplification of inherited biases via fine-tuning techniques \cite{shi2023towards, zhuang2023foundation}. In particular, the under-represented groups might be subjected to inequities in treatment. This calls for the exploration of group fairness techniques robust against biases \cite{zhang2024unified, salazar2024survey}. Existing works mainly achieve the FL group fairness regarding a single sensitive attribute from three perspectives: aggregation, reweighting and client participation \cite{kanaparthy2022fair, zeng2024fair, du2021fairness, ezzeldin2023fairfed, zhang2020fairfl, salazar2023fair}. Notably, Zeng et al. \cite{zeng2024fair} designed a fairness-aware adaptation framework FF-DVP to mitigate the bias of pre-trained vision-language models (VLMs) and learn fairness constraints by aggregating modality-fused classification heads. As a well-known reweighting method, Fairfed \cite{ezzeldin2023fairfed} adaptively adjusted local weights on the server side by calculating the deviation between the global and local fairness metrics. FAIR-FATE \cite{salazar2023fair} prioritizes clients with higher fairness to participate in the federation by estimating local model contributions using a Momentum term. In practice, datasets often consist of multiple sensitive attributes. Although current research \cite{meerza2024glocalfair, djebrouni2024bias} has optimized the model utility under different sensitive attributes via fairness constraints, they are insufficient in the interpretability of the dependencies between sensitive attributes and FFM group fairness, thereby hurting transparency and trust in sensitive applications.

A comprehensive understanding of the relationship between sensitive attributes and group fairness can be constructed from two complementary perspectives: 1) the correlations between attributes and labels, and 2) the intersections of multiple sensitive attributes \cite{dorleon2022feature, poulain2023improving, wang2024analyzing}. The former (e.g., feature selection) indicates potential biases affecting fairness via a strong correlation between a sensitive attribute and an adverse outcome \cite{dorleon2022feature}, while the latter focuses on understanding how the intersection of multiple sensitive attributes impacts fairness outcomes \cite{poulain2023improving, wang2024analyzing}. However, correlation alone cannot reveal whether these relationships are causal or simply spurious associations caused by confounding factors. Causal analysis can reveal how sensitive attributes influence fairness outcomes directly or indirectly, allowing for the explicit modeling of causal pathways and identifying root causes of biases \cite{guo2024fedcsl, li2024causal, feuerriegel2024causal}. Specifically, causal discovery seeks to uncover the hidden cause-and-effect relationships among variables by defining the causal structure and constructing a causal graph \cite{guo2024fedcsl}. Causal inference can quantify the causal effects of interventions, enabling precise predictions of how addressing bias for one attribute might impact other attributes or demographic decisions \cite{li2024causal, feuerriegel2024causal}. This study is a first-of-its-kind attempt to leverage causal frameworks to explore rigorous and interpretable approaches for balancing FFM group fairness with multiple sensitive attributes.

\section{Problem Definition}
\label{sec:prob}

In our study, we extend FF-DVP \cite{zeng2024fair}, which is the first approach for achieving group fairness in the field of FFM. We refer to the extended version of FF-DVP as \underline{E}xtended \underline{FF-DVP} (\emph{\methodname{}}). To adapt to multiple sensitive attributes, we reconstruct the local VLM architecture and redefine the key problem formulas. As shown in Figure \ref{fig:overview}, the local model of \methodname{} fine-tunes the large-scale FFM by solely updating and sharing the important orange modules. Various demographic-only prompts converted by sensitive attributes input the text encoder in parallel to dynamically trade off the group fairness. Local clients also analyze the causal effect to interpret the relationship between sensitive attributes and group fairness.

Suppose there are $\mathcal{S}=\{1,2,..., S\}$ clients and a server (indexed as $g$). We denote the local dataset at client $s$ as $\mathcal{D}_s=\{(x,a,y) \vert x\in\mathcal{X}, a\in\mathcal{A}, y \in \mathcal{Y}\}$, where $\mathcal{X}$ and $\mathcal{Y}$ are the input feature and the label space, respectively. The local sensitive attribute set $\mathcal{A}=\{A^1, A^2, ..., A^K\}$ has $K$ sensitive attributes. There are $\mathcal{T}=\{1,2,...,T\}$ communication rounds and each communication round runs $\mathcal{E}=\{1,2,...,E\}$ local training epochs.

\subsection{Local-level Fairness} 

Each client equalizes relevance between demographic-only text prompts and images to debias the local VLM by generating a local-level fairness regularizer $\mathcal{L}_{lf}$. Multiple demographic-only prompts are allowed to input the network in parallel with images and encoded to independent text representations, which helps further analyze the respective causal effect between sensitive attributes with labels. For $k$-th sensitive attribute $A^k$, $\mathcal{L}_{lf}^k$ is derived by the KL-divergence to maximize the relevance equilibrium of demographic-only text representations and visual representations:
\begin{equation}
    Pr(A^k)=Softmax(\frac{cos(\tilde{z},t_{a_1})}{\tau}, ..., \frac{cos(\tilde{z},t_{a_{|A^k|}})}{\tau}),
\end{equation}
\begin{equation}
    \mathcal{U}(1,|A^k|)=[\frac{1}{|A^k|}, ..., \frac{1}{|A^k|}],
\end{equation}
\begin{equation}
    \mathcal{L}_{lf}^k = KL(Pr(A^k)||\mathcal{U}(1,|A^k|)),
\end{equation}
where $Pr(A^k)$ computes the normalized cosine similarities of demographic-only text representations $t_{a}$ and visual representations $\tilde{z}$. $\mathcal{U}(1,|A^k|)$ represents a uniform distribution. Thus, this equation aims to avoid images being biased towards any demographic group. 

To dynamically trade off the bias effect of various sensitive attributes, we combine $\mathcal{L}_{lf}^k$ through weighted averaging to formulate the local-level fairness regularizer $\mathcal{L}_{lf}$.
\begin{equation}
    \mathcal{L}_{lf} = \sum_{k\in K} \alpha^k \mathcal{L}_{lf}^k,
\end{equation}
where $\alpha^k$ is the weight and $\sum_{k\in K} \alpha^k=1$.

\subsection{Global-level Fairness} 

A modality-fused classifier is designed to predict the decision $\hat{\mathcal{Y}}$ by taking the visual representations $\tilde{z}$ and ground-truth text representations $t_{gt}$ as inputs. To control the group fairness from the final prediction outcomes, clients also deduce global-level fairness regularizer $\mathcal{L}_{gf}$ via the traditional fairness notions $\Phi$, such as demographic parity $\Phi_{DP}$ and equalized odds $\Phi_{EO}$. 

For the $k$-th sensitive attribute $A^k$, demographic parity $\Phi_{DP}^k$ quantifies the bias of the probability of a positive outcome across different demographic groups:
\begin{equation} \label{eq:DP}
    \Phi_{DP}^k = |P(\hat{\mathcal{Y}}=1|A^k=0)-P(\hat{\mathcal{Y}}=1|A^k=1)|.
\end{equation}

Equalized odds $\Phi_{EO}^k$ is a stricter fairness metric that measures the difference in the true positive and false positive rates between two groups, conditional on the true outcome:
\begin{equation} \label{eq:EO}
\begin{aligned}
    \Phi_{EO}^k = &|P(\hat{\mathcal{Y}}=1|\mathcal{Y}=y,A^k=0) \\
    & -P(\hat{\mathcal{Y}}=1|\mathcal{Y}=y,A^k=1)|, \forall y \in \{0,1\}.
\end{aligned}
\end{equation}

The global-level fairness regularizer $\mathcal{L}_{gf}$ is the similar weighted average of $\Phi^k$:
\begin{equation}
    \mathcal{L}_{gf} = \sum_{k\in K} \beta^k \Phi^k,
\end{equation}
where $\beta^k$ is the weight and $\sum_{k\in K} \beta^k=1$.

In addition to the above fairness regularizers, we also incorporate a contrastive loss regularization $\mathcal{L}_{con}$ to improve the prediction accuracy by minimizing the distance between the visual representations $\tilde{z}$ and ground-truth text representations $t_{gt}$. These three regularizers act on the supervised loss (i.e., cross-entropy loss) together to correct the optimization direction of local models. For fine-tuning FFMs with massive parameters, local clients share prompted image encoders and modality-fused classifiers with the server to improve the robustness and communication efficiency, where the server aggregates the local trainable parameters $W_s$ via the weighted average ($W = \sum_s \frac{|\mathcal{D}_s|}{\sum_s |\mathcal{D}_s|} W_s, \forall s \in \mathcal{S}$).
Since this part is not associated with fairness, we keep it the same as under FF-DVP and omit detailed descriptions due to space limitations.

\section{Causal Analysis}
\label{sec:caus}

The structured approach, grounded in causal discovery and inference, provides a robust method for advancing fairness-aware machine learning, particularly in complex federated settings where traditional statistical methods fall short \cite{binkyte2023causal}.

\subsection{Causal Discovery} 

To analyze the relationship between sensitive attributes and group fairness, it is essential to first understand the causal structure that connects these variables. Causal discovery provides a systematic approach to uncovering this structure directly from data, whose goal is to construct a causal directed acyclic graph (DAG) $\mathcal{G}=(V,E)$, where $V$ are the variables of interest and $E$ are the causal relationships between them. For instance, the presence of a direct causal link between $k$-th sensitive attribute $A^k$ and the label $\mathcal{Y}$: $A^k \rightarrow \mathcal{Y}$ would imply that $A^k \not \! \perp \!\!\! \perp \mathcal{Y}|Z$, where $Z \subseteq \{\mathcal{X}, \mathcal{A}\} \setminus A^k$ and $\not \! \perp \!\!\! \perp$ represents the dependence relation. If $A^k$ is conditionally independent for $\mathcal{Y}$ (i.e., $A^k\perp \!\!\! \perp \mathcal{Y}|Z$), the edge between $A^k$ and $\mathcal{Y}$ will be deleted. Using algorithms like PC, GES, or Functional Causal Models (e.g., LiNGAM), the relationships among variables are inferred through conditional independence testing.
\begin{equation}
    A^k\perp \!\!\! \perp \mathcal{Y}|Z \Longleftrightarrow P(A^k, \mathcal{Y}|Z) = P(A^k|Z) \cdot P(\mathcal{Y}|Z).
\end{equation}

\subsection{Causal Inference} 

Once the causal graph $\mathcal{G}$ is constructed, causal inference methods can be employed to quantify the strength of the relationships captured within the graph through the causal effect estimation. A key quantity of effect is the total effect (TE), which is defined as the difference in the expected value when the sensitive attribute $A^k$ is intervened upon to take two different values, effectively removing any confounding influence.
\begin{equation} \label{eq:TE}
    TE_{A^k \rightarrow \mathcal{Y}} = \mathbb{E}[\mathcal{Y}|do(A^k=0)] - \mathbb{E}[\mathcal{Y}|do(A^k=1)],
\end{equation}
where $\mathbb{E}[\mathcal{Y}]=\sum_{y} y \cdot P(\mathcal{Y}=y)$ for discrete data.

TE can be decomposed into two components: natural direct effect (NDE) and natural indirect effect (NIE), i.e., $TE=NDE+DIE$. NDE captures the impact of $A^k$ on $\mathcal{Y}$ through causal pathways that do not involve other mediating variables $M$ (e.g., $A^k \rightarrow \mathcal{Y}$), isolating the influence of $A^k$ while holding other intermediate variables constant.
\begin{equation} \label{eq:NDE}
\begin{aligned}
    NDE_{A^k \rightarrow \mathcal{Y}} = &\mathbb{E}[\mathcal{Y}|do(A^k=0), M=m] \\
    &- \mathbb{E}[\mathcal{Y}|do(A^k=1), M=m].
\end{aligned}
\end{equation}
NIE quantifies the influence of $A^k$ on $\mathcal{Y}$ that operates through mediating variables $M$ (i.e., $A^k \rightarrow M \rightarrow \mathcal{Y}$). It highlights how systemic biases in mediators propagate the influence of sensitive attributes on the outcome.
\begin{equation} \label{eq:NIE}
\begin{aligned}
    NIE_{A^k \rightarrow \mathcal{Y}} = &\mathbb{E}[\mathcal{Y}|do(A^k=1), M=do(M(A^k=0))] \\
    &- \mathbb{E}[\mathcal{Y}|do(A^k=1), M=do(M(A^k=1))].
\end{aligned}
\end{equation}

\subsection{Relationship with Group Fairness} 

Understanding how causal effects relate to group fairness provides a deeper theoretical foundation for fairness interventions, where group fairness metrics can be expressed in terms of the causal pathways identified in the graph. For example, we can observe from Eq. (\ref{eq:DP}), Eq. (\ref{eq:EO}), Eq. (\ref{eq:TE}) - Eq. (\ref{eq:NIE}) that demographic parity corresponds to eliminating the causal effect of sensitive attributes on model predictions.
A large causal effect value implies that the sensitive attribute has a significant influence on the label, indicating potential bias in the FFM. 
\begin{equation} \label{eq:relation}
    |TE_{A^k \rightarrow \mathcal{Y}}| \propto \frac{1}{|\Delta\Phi^k|},
\end{equation}
where $\Delta\Phi^k=\Phi^k(unbias)- \Phi^k(bias)$. $\Phi^k(bias)$ is the fairness with inherent bias and $\Phi^k(unbias)$ is the fairness after mitigating bias.

In the context of FL, causal analysis ensures that the graph integrates heterogeneous data distributions from different clients via parameter aggregation, which is crucial for capturing global causal relationships without bias from local data. In addition, causal perspective enhances the robustness and interoperability of fairness strategies and aligns with ethical principles, rather than merely addressing surface-level correlations.

\section{Experimental Evaluation}
\label{sec:exper}

\subsection{Dataset}

Following \cite{zeng2024fair}, we conduct experiments on two public face datasets: CelebA and FairFace. CelebA contains over 200,000 celebrity face images, each annotated with 40 binary attribute labels (e.g., gender, smiling, eyeglasses). We choose \emph{gender} and \emph{age} as sensitive attributes to assist images for predicting labels \emph{attractive} or \emph{smiling}. The FairFace dataset is a balanced face image dataset designed to address biases in facial attribute recognition tasks, which includes over 108,000 images with three attributes (gender,  seven racial categories, and a broad age range). \emph{Gender} and \emph{race} are chosen as sensitive attributes to predict the label \emph{age}. In FL settings, the server randomly samples 20\% instances as the testing set and others are divided into 5 clients with a local training and validation ratio of 4:1. 

\subsection{Experiment Settings}

Our experiments use the CLIP model with the configuration of ViT-L/14@336px to encode the images and text prompts, where the prompt with the highest normalized cosine similarity to the image is viewed as the predicted outcome. The modality-fused classifier is composed of a two-layer fully connected network.  We choose the AdamW optimizer with learning rates of \{1e-5, 5e-4\}. The global communication round is set to 2 or 4 based on the convergence state and each round runs 2 local epoches.

\subsection{Evaluation Metrics}

We evaluate \methodname{} based on the following metrics:

\textbf{Accuracy} Balanced accuracy (Acc) is the average accuracy for all demographic groups under multiple sensitive attributes:
\begin{equation}
    Acc = \frac{ \sum_{A^k} \sum_a \sum_y P(\hat{\mathcal{Y}}=\mathcal{Y}|A^k=a) }{|\mathcal{A}| \cdot |A^k| \cdot |\mathcal{Y}|}, \forall A^k \in \mathcal{A}.
\end{equation}
Accuracy parity $\Phi_{AP}$ is the sum of absolute errors among all demographic groups under multiple sensitive attributes:
\begin{equation}
\begin{aligned}
    \Phi_{AP} = &\frac{1}{|\mathcal{A}|} \sum_{A^k} \left [ \sum_a \sum_y \sum_{a'} \sum_{y'} | P(\hat{\mathcal{Y}}=y|A^k=a) \right.\\
    &\left. - P(\hat{\mathcal{Y}}=y'|A^k=a') | \right ], \forall A^k \in \mathcal{A}.
\end{aligned}
\end{equation}

\textbf{Fairness} The group fairness is measured by demographic parity $\Phi_{DP}^k$ and equalized odds $\Phi_{EO}^k$ as in Eq. (\ref{eq:DP}) and Eq.(\ref{eq:EO}) presented. $\Delta \Phi^k$ in Eq. (\ref{eq:relation}) reveals the effect of removing bias.

\textbf{Causal Effect} Based on Eq. (\ref{eq:TE}), we compute the causal effect $TE_{A^k \rightarrow \mathcal{Y}}$ (abbreviated as $TE^k$) between the sensitive attribute $A^k$ and ground-truth label $\mathcal{Y}$ to assess the real causal relationship of variables themselves. p-value measures whether the observed difference between the original and new estimates is statistically significant. 

\begin{table*}[ht]
\renewcommand{\arraystretch}{1.1}
\caption{Results about Accuracy and Group Fairness under different sensitive attributes.}
\label{tab:acc_fair}
\centering
\resizebox{1\textwidth}{!}{
    \begin{tabular}{|ccc|ccccccrrrr|}
    \hline
     \makecell*[c]{Task (Dataset)} & $A^1$ & $A^2$ & Acc  & $\Phi_{AP}$  & $\Phi_{DP}^1$ & $\Phi_{EO}^1$  & $\Phi_{DP}^2$  & $\Phi_{EO}^2$  & $\Delta\Phi_{DP}^1$ & $\Delta\Phi_{EO}^1$  & $\Delta\Phi_{DP}^2$  & $\Delta\Phi_{EO}^2$ \\ 
    \hline
    \multirow{4}{*}{\makecell{Attractive Detection \\ (CelebA)}}  &  -  &  -  &  \textbf{0.758}   &  1.138   &  0.138   &  0.277   &  0.478    &  0.542   &  \makecell[c]{-}   &  \makecell[c]{-}   &  \makecell[c]{-}    &  \makecell[c]{-}   \\ 
      &  $\surd$  &  -  &  0.755   &  0.866   &  \textbf{0.049}   &  \textbf{0.110}   &  0.491   &  0.542   &  \textbf{-0.089}    &  \textbf{-0.167}    &  0.013     &  0.000    \\
      &  -  &  $\surd$  &  0.742   &  1.135   &  0.139  &  0.278   &  \textbf{0.450}   &  \textbf{0.531}   &  0.001    &  0.001    &  \textbf{-0.028}     &  \textbf{-0.011}    \\
      &  $\surd$  &  $\surd$  &  0.727   &  \textbf{0.755}   &  0.112   &  0.223   &  0.468   &  0.535   &  -0.026    &  -0.054    &  -0.010     &  -0.007   \\ 
    \hline
    \multirow{4}{*}{\makecell{Smiling Detection \\ (CelebA)}} &  -  &  -  &  \textbf{0.879}   &  0.380  &  0.193   &  0.386   &  0.072   &  0.074   &  \makecell[c]{-}   &  \makecell[c]{-}   &  \makecell[c]{-}    &  \makecell[c]{-}   \\
      &  $\surd$  &  -  &  0.867  &  0.346   &  \textbf{0.004}  &  \textbf{0.019}   &  0.105   &  0.081   &  \textbf{-0.189}    &  \textbf{-0.367}    &  0.033     &  0.007    \\
       &  -  &  $\surd$  &  0.854  &  1.398   &   0.264   &  0.527   &  \textbf{0.063}   &   \textbf{0.008}   &  0.071    &  0.141    &  -0.009     &  \textbf{-0.066}   \\ 
       &  $\surd$  &  $\surd$  &  0.841   &  \textbf{0.153}   &   0.131   &  0.263   &  0.052   &  0.069   &  -0.062    &  -0.123    &  \textbf{-0.020}     &  -0.005    \\ 
    \hline
    \multirow{4}{*}{\makecell{Age Detection \\ (FairFace)}} &  -  &  -  &  \textbf{0.864}   &  0.181   &  0.068   &  0.127   &  0.049  &  0.038   &  \makecell[c]{-}   &  \makecell[c]{-}   &  \makecell[c]{-}    &  \makecell[c]{-}   \\
      &  $\surd$  &  -  &  0.855   &  0.416   &  0.060   &  0.121   &  0.063   &  0.067   &  -0.008    &  -0.006    &  0.014     &  0.029    \\
      &  -  &  $\surd$  &  0.863    &  \textbf{0.247}   &  0.070   &  0.139   &  \textbf{0.019}  &  \textbf{0.021}   &  0.002    &  0.012    &  \textbf{-0.030}     &  \textbf{-0.017}    \\
       &  $\surd$  &  $\surd$  &  0.856   &  0.320   &  \textbf{0.030}   &  \textbf{0.059}   &  0.042   &  0.025   &  \textbf{-0.038}    &  \textbf{-0.068}    &  -0.007     &  -0.013    \\ 
    \hline
    \end{tabular}
}
\end{table*}

\begin{table*}[ht]
\renewcommand{\arraystretch}{1.2}
\caption{Results about causal analysis and robustness evaluation of causal estimate.}
\label{tab:causal}
\centering
\resizebox*{1\linewidth}{!}{
    \begin{tabular}{|cc|cccccccccc|cc|}
    \hline 
    \multirow{2}{*}{Task (Dataset)} & \multirow{2}{*}{Refute} & \multicolumn{2}{c}{Client 1} & \multicolumn{2}{c}{Client 2} & \multicolumn{2}{c}{Client 3} & \multicolumn{2}{c}{Client 4} & \multicolumn{2}{c}{Client 5} & \multicolumn{2}{|c|}{Average} \\ 
    \cline{3-14} 
    \rule{0pt}{9pt} & & $TE^1$  & $TE^2$  & $TE^1$  & $TE^2$  & $TE^1$  & $TE^2$  & $TE^1$  & $TE^2$  & $TE^1$  & $TE^2$  & $TE^1$  & $TE^2$ \\ 
    \hline
    \multirow{3}{*}{\makecell{Attractive Detection \\ (CelebA)}}  &  Old  &  -0.122     &  -0.112     &  0.536     &  0.565     &  -0.240      &  0.637     &  -0.215     &  0.486     & 0.319     &  0.390     &  0.055      &  \textbf{0.393}     \\ 
      &  New  &  -0.121    &  -0.114    &  0.526    &  0.564    &  -0.236    &  0.632     &  -0.208    &  0.461    &  0.318    &  0.378    &  0.056      &  \textbf{0.384}     \\
      &  p-value  &  0.920    &  0.900    &  0.980    &  0.960    &  0.940   &  0.940     &  0.980    &  0.860    &  0.880    &  0.940    &  \textbf{0.940}     &  0.920     \\
    \hline
    \multirow{3}{*}{\makecell{Smiling Detection \\ (CelebA)}} &   Old  &  0.394   &  0.021   &  -0.604   &  0.017   &  -0.021    &  -0.508   &  -0.227   &  -0.154   &  0.559   &  0.831   &  0.020    &  \textbf{0.041}   \\ 
      &  New  &  0.392   &  0.029   &  -0.624   &  0.014   &  -0.024   &  -0.493   &  -0.240   &  -0.152   &  0.568   &  0.818   &  0.015    &  \textbf{0.043}   \\
      &  p-value  &  0.980     &  0.900     &  0.940    &  0.980    &  1.000   &  0.960     &  0.920    &  0.940    &  0.820    &  0.960    &  0.932     &  \textbf{0.948}     \\
    \hline
    \multirow{3}{*}{\makecell{Age Detection \\ (FairFace)}} &   Old  &  -0.433     &  0.047     &  -0.142     &  -0.146     &  -0.362      &  -0.032     &  0.205     &  0.240     &  0.289     &  0.215     &  -0.088      &  \textbf{0.065}     \\ 
      &  New  &  -0.448    &  0.029    &  -0.161    &  -0.156    &  -0.374    &  -0.033    &  0.208    &  0.239    &  0.292    &  0.215    &  -0.097      &  \textbf{0.059}     \\
      &  p-value  &  0.800    &  0.900    &  0.960    &  0.880    &  0.840   &  0.900     &  0.960    &  0.960    &  0.940    &  0.900    &  0.900     &  \textbf{0.908}     \\
    \hline
    \end{tabular}
}
\end{table*}

\subsection{Results about Accuracy and Group Fairness}

In our study, attributes $A^1$ and $A^2$ represent \emph{gender} and \emph{age} for the CelebA dataset and \emph{gender} and \emph{race} for the FairFace dataset, respectively. The check mark $\surd$ in table \ref{tab:acc_fair} is the signal that this experiment mitigates the bias of the corresponding attribute. From the results in table \ref{tab:acc_fair}, we observed that accuracy and group fairness are hard to trade off or improve together. Enhancing the group fairness is no doubt to degrade the accuracy to some extent. In addition, different attributes will `compete' with each other where the alleviation of one attribute bias will lead to an increase in another attribute bias. For example, in the smiling detection task, mitigating the gender ($A^1$) bias will decrease the demographic parity of gender $\Phi_{DP}^1$ by 0.189 ($\Delta\Phi_{DP}^1$) while increasing the demographic parity of age by 0.033 ($\Delta\Phi_{DP}^2$).

\subsection{Results about Causal Analysis}

To validate the relationship between sensitive attributes and group fairness, we quantified the causal effect among sensitive attributes and labels across all clients of the FL system. In addition, we supplement the 'refute' experiments to test the robustness of a causal estimate by introducing synthetic random common causes (confounders) into the causal graph. Results in table \ref{tab:causal} show that each client has different causal effects due to the imbalanced data distribution. On average, gender has a weaker causal effect on the label than other attributes. For example, being male increases the likelihood of attractiveness by 0.055, which is lower than being a young person with an increase of 0.393. Thus, the decision of attractiveness performs less dependence on the gender and the bias on the gender will be eliminated easier, which is supported by results in tabel \ref{tab:acc_fair} that the optimal $\Delta\Phi_{DP}^1$ (-0.089) is lower than $\Delta\Phi_{DP}^2$ (-0.028). Moreover, the small gap between old and new causal effects and high p-value suggest that the addition of random confounders does not significantly affect the estimate, indicating the robustness of our causal effect estimation.

\section{Conclusions and Future Work}

In this article, we focus on improving the interpretability of group fairness in FFM systems from the perspective of causal analysis. Under realistic settings involving multiple sensitive attributes, we propose to extend FF-DVP to balance the group fairness with different sensitive attributes dynamically. In addition, causal discovery and inference help to construct the causal graph among variables and quantify the causal effect, which reveals the underlying correlation between sensitive attributes and group fairness in theory. Experiments have demonstrated that the larger the causal effect, the harder it is to ensure group fairness.

From our findings, it can be concluded that developing techniques of adaptive fairness-aware algorithms to mitigate biases across multiple sensitive attributes while maintaining high model performance, is an ongoing challenge in this area. In the long run, integrating interpretability mechanisms into these real-world systems is essential for practical adoption, which helps domain experts validate fairness outcomes, adjust detected biases and build trust in the system. In subsequent research, we plan to develop FFM techniques that can achieve group fairness in the face of multiple sensitive attributes with good interpretability.

\bibliographystyle{IEEEbib}
\bibliography{ref}

\end{document}